\documentclass[letterpaper]{article}
\usepackage{aaai20}
\usepackage{times}
\usepackage{helvet}
\usepackage{courier}
\usepackage[hyphens]{url} 
\usepackage{graphicx}
\usepackage{graphicx}
\usepackage{subcaption}
\usepackage{multirow}

\usepackage{enumitem}\setlist{nolistsep,leftmargin=1.5em}

\newcommand{\citeauthoryearp}[1]{\citeauthor{#1}~(\citeyear{#1})}

\title{Game Level Clustering and Generation using Gaussian Mixture VAEs}
\author{
Zhihan Yang\textsuperscript{1}, Anurag Sarkar\textsuperscript{2} and Seth Cooper\textsuperscript{2} \\
\textsuperscript{1}Carleton College\\
\textsuperscript{2}Northeastern University\\ 
yangz2@carleton.edu,
sarkar.an@northeastern.edu,
se.cooper@northeastern.edu\\
}

\begin{document}
\maketitle
\begin{abstract}
Variational autoencoders (VAEs) have been shown to be able to generate game levels but require manual exploration of the learned latent space to generate outputs with desired attributes. While conditional VAEs address this by allowing generation to be conditioned on labels, such labels have to be provided during training and thus require prior knowledge which may not always be available. In this paper, we apply Gaussian Mixture VAEs (GMVAEs), a variant of the VAE which imposes a mixture of Gaussians (GM) on the latent space, unlike regular VAEs which impose a unimodal Gaussian. This allows GMVAEs to cluster levels in an unsupervised manner using the components of the GM and then generate new levels using the learned components. We demonstrate our approach with levels from \textit{Super Mario Bros.}, \textit{Kid Icarus} and \textit{Mega Man}. Our results show that the learned components discover and cluster level structures and patterns and can be used to generate levels with desired characteristics.
\end{abstract}

%%%%%%%%%%%%%%%%%%%%%%%%%%%%%%%%%%%%%%%%%%%%%%%%%%
%% FIGURE/TABLE BEGIN %%%%%%%%%%%%%%%%%%%%%%%%%%%%
%%%%%%%%%%%%%%%%%%%%%%%%%%%%%%%%%%%%%%%%%%%%%%%%%%

%================================================================================
% FIGURES

\newcommand{\XFIGUREgmvae}{
\begin{figure}[t]
\centering
\includegraphics[width=1\columnwidth]{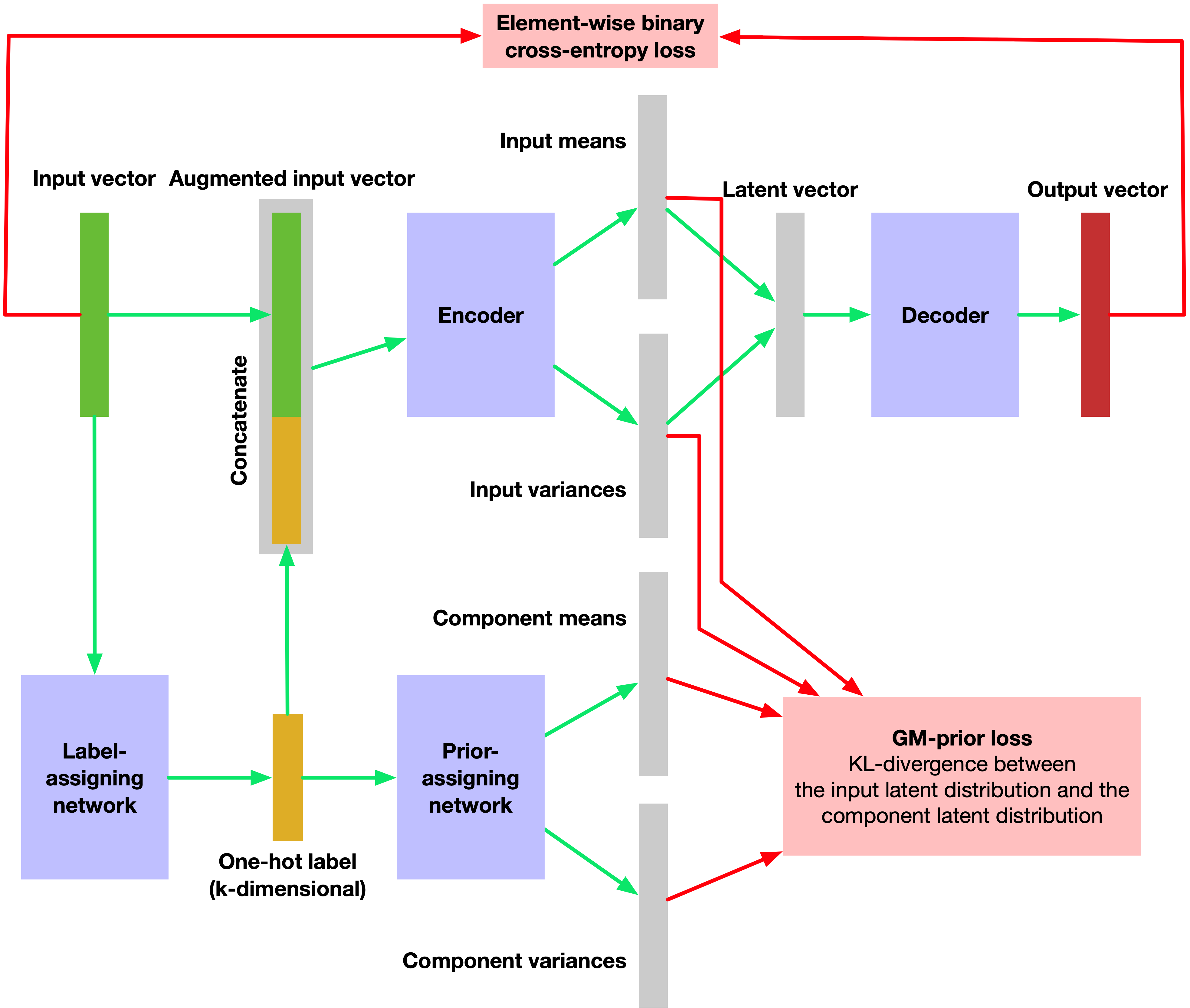}
\caption{A GMVAE with a $k$-component GM prior.}
\label{fig:gmvae}
\end{figure}
}

\newcommand{\XFIGUREdisentanglement}{
\begin{figure*}[!th]
\centering
%\resizebox{\textwidth}{!}{%
\begin{tabular}{cccc}
& SMB & KI & MM 
\\ 
\rotatebox{90}{\scriptsize{\hspace{35pt}\textbf{GMVAE}}} &
\includegraphics[width=0.275\textwidth]{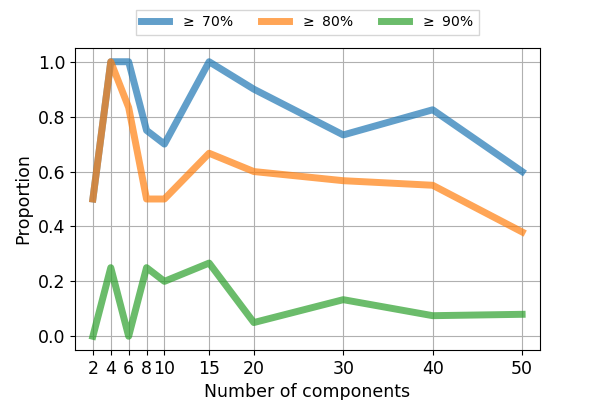} & 
\includegraphics[width=0.275\textwidth]{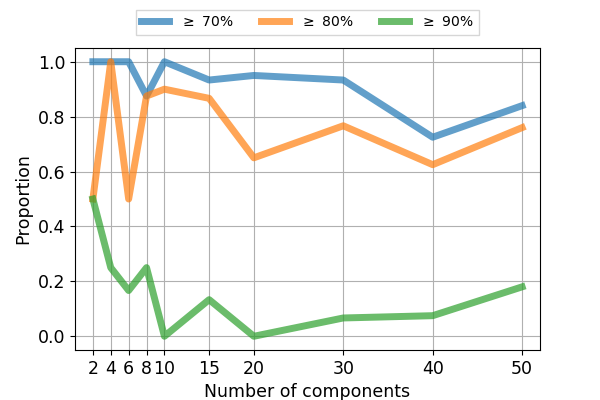} &
\includegraphics[width=0.275\textwidth]{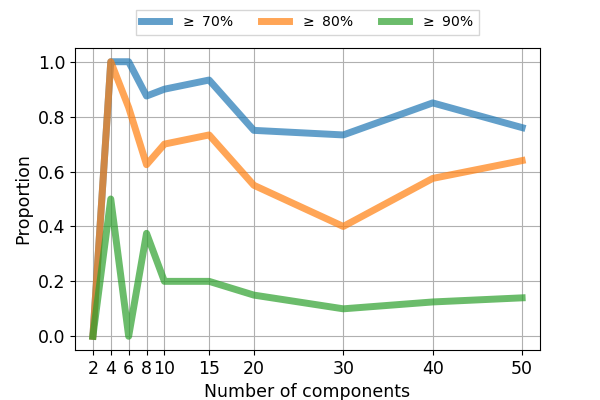}
\\
\rotatebox{90}{\scriptsize{\hspace{30pt}\textbf{VAE-GMM}}} &
\includegraphics[width=0.275\textwidth]{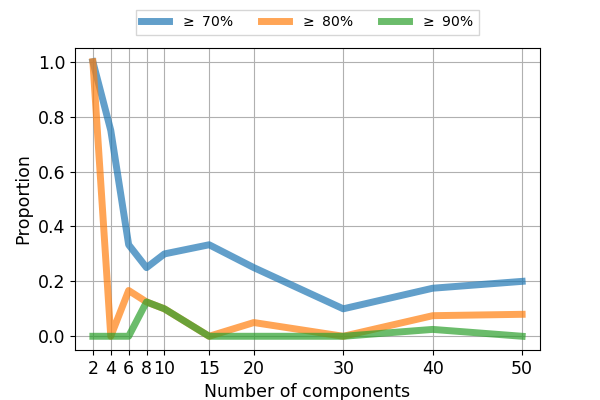} &
\includegraphics[width=0.275\textwidth]{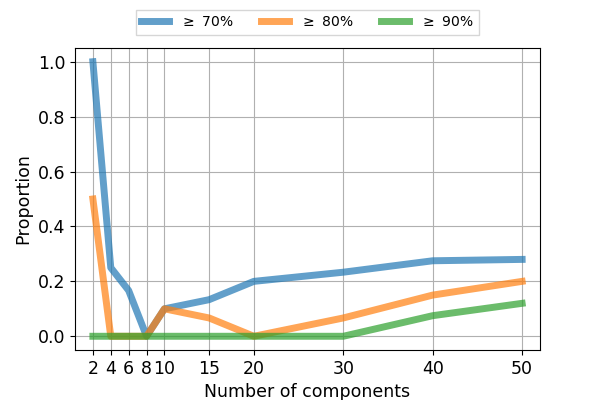} &
\includegraphics[width=0.275\textwidth]{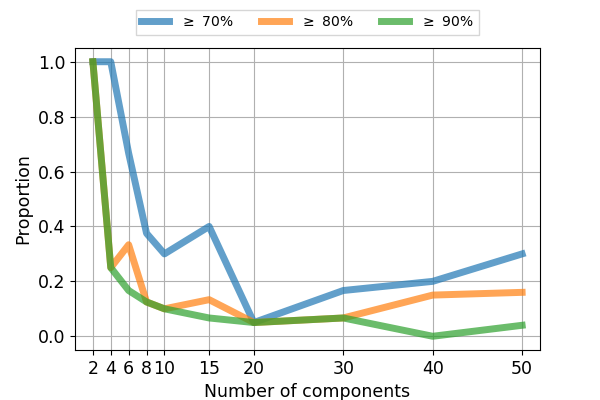} 
\end{tabular}
%}
\caption{Proportion of components being ${\geq} 70\%$ (mediocre), ${\geq}80\%$ (good) and ${\geq} 90\%$ (excellent) disentangled against the number of components.}
\label{fig:disentanglement}
\end{figure*}
}

\newcommand{\XFIGURElatentspacevis}{
\begin{figure}[!th]
    \begin{subfigure}[h]{0.475\columnwidth}
        \centering
        \includegraphics[width=\textwidth]{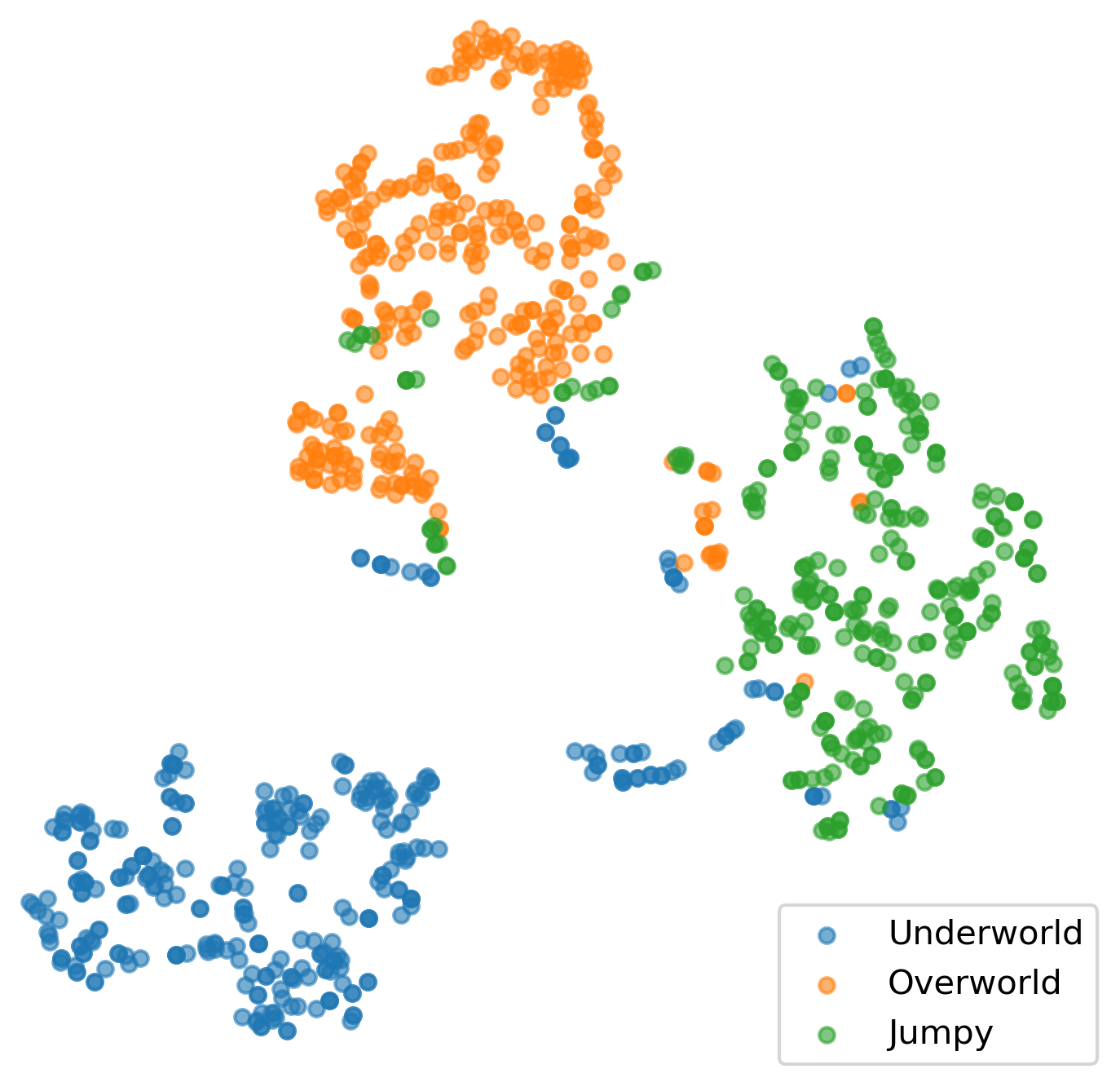}
        \captionsetup{justification=centering}
        \caption{GMVAE latent space color-coded using level-type labels}
        \label{fig:gmvae latent space label}
    \end{subfigure}
    \begin{subfigure}[h]{0.475\columnwidth}  
        \centering 
        \includegraphics[width=\textwidth]{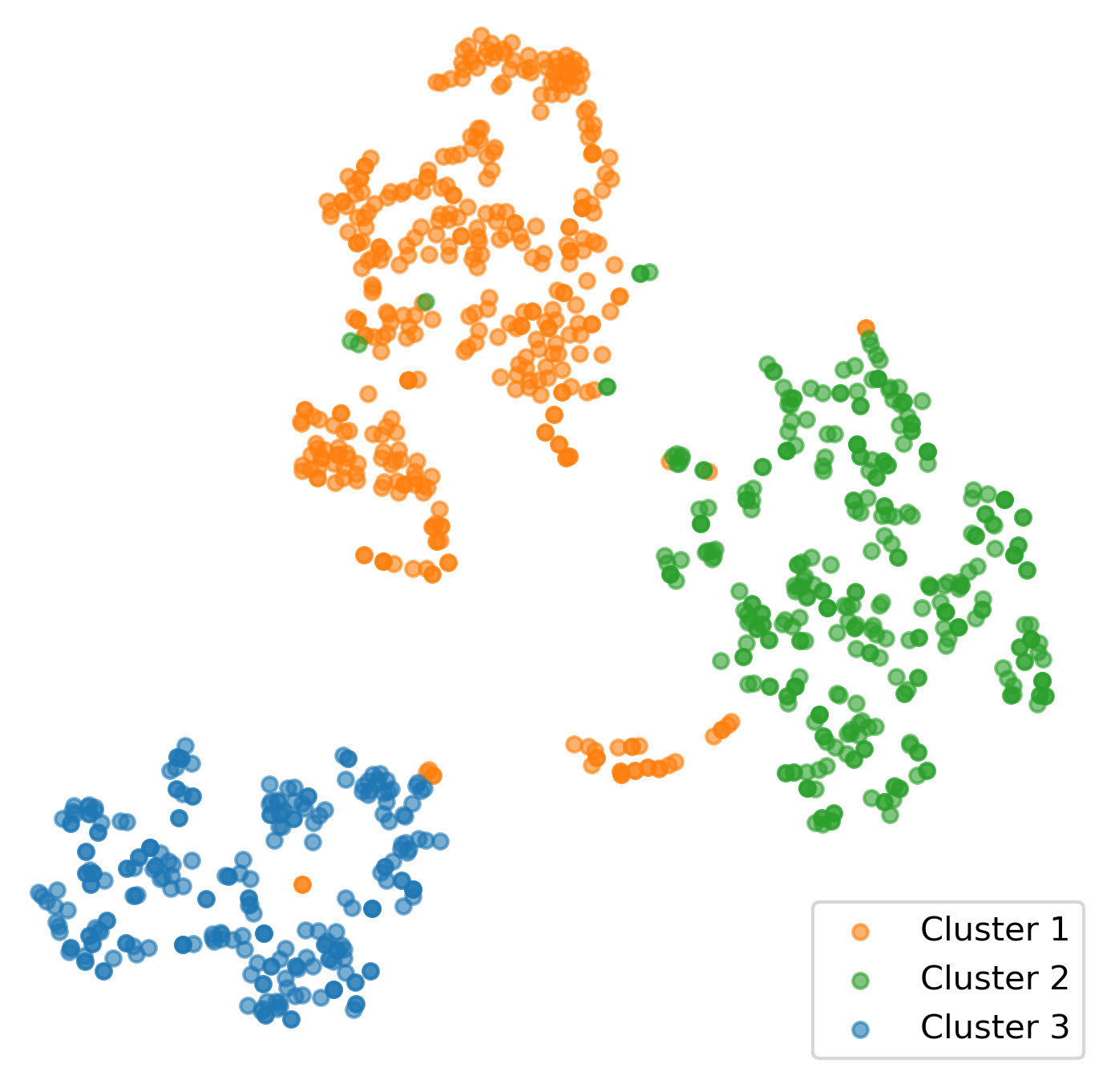}
        \captionsetup{justification=centering}
        \caption{GMVAE latent space color-coded using learned clusters}
        \label{fig:gmvae latent space cluster}
    \end{subfigure}
    \begin{subfigure}[h]{0.475\columnwidth}   
        \centering 
        \includegraphics[width=\textwidth]{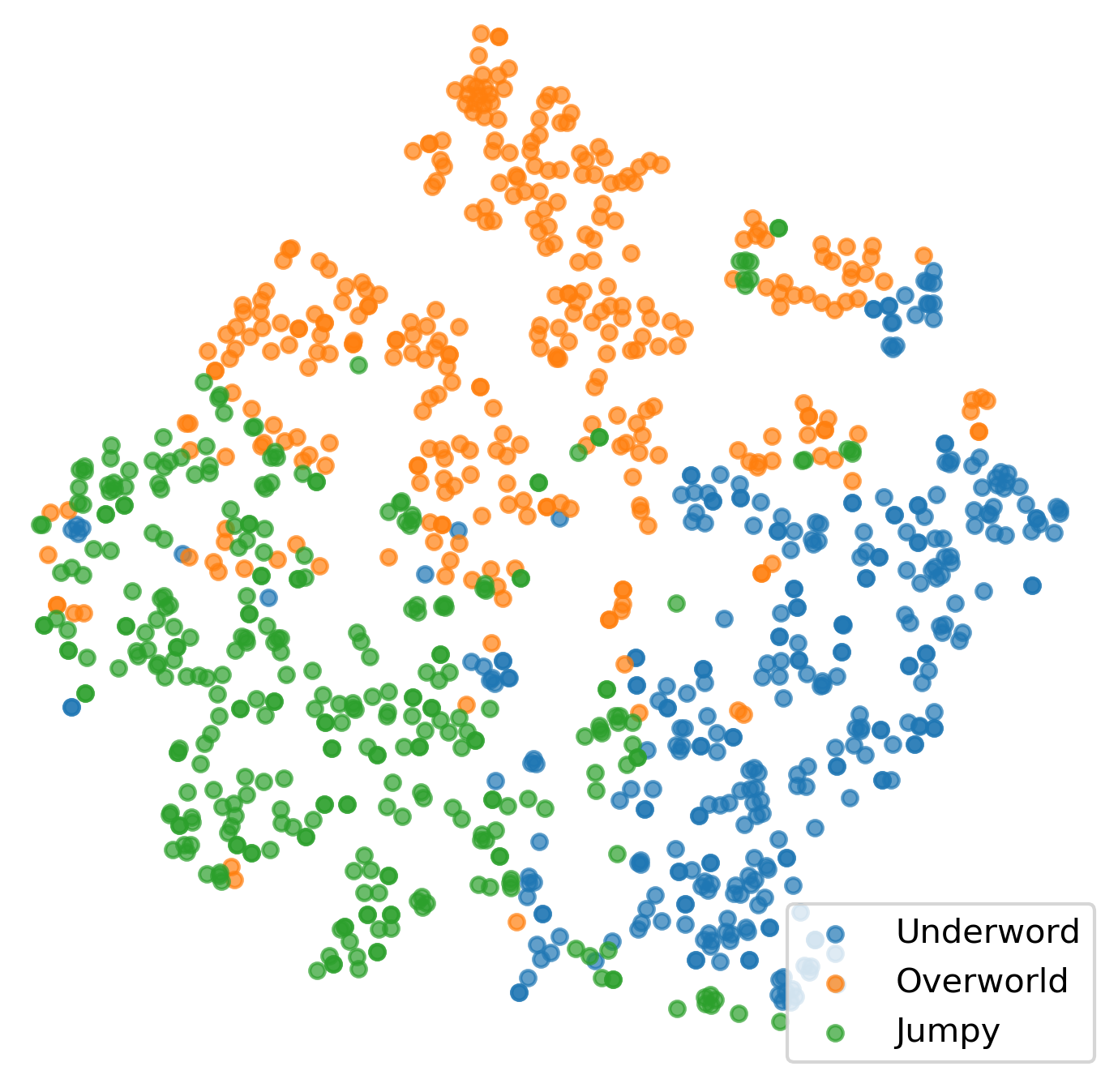}
        \captionsetup{justification=centering}
        \caption{VAE-GMM latent space color-coded using level-type labels}  
        \label{fig:vaegmm latent space label}
    \end{subfigure}
    \quad
    \begin{subfigure}[h]{0.475\columnwidth}   
        \centering 
        \includegraphics[width=\textwidth]{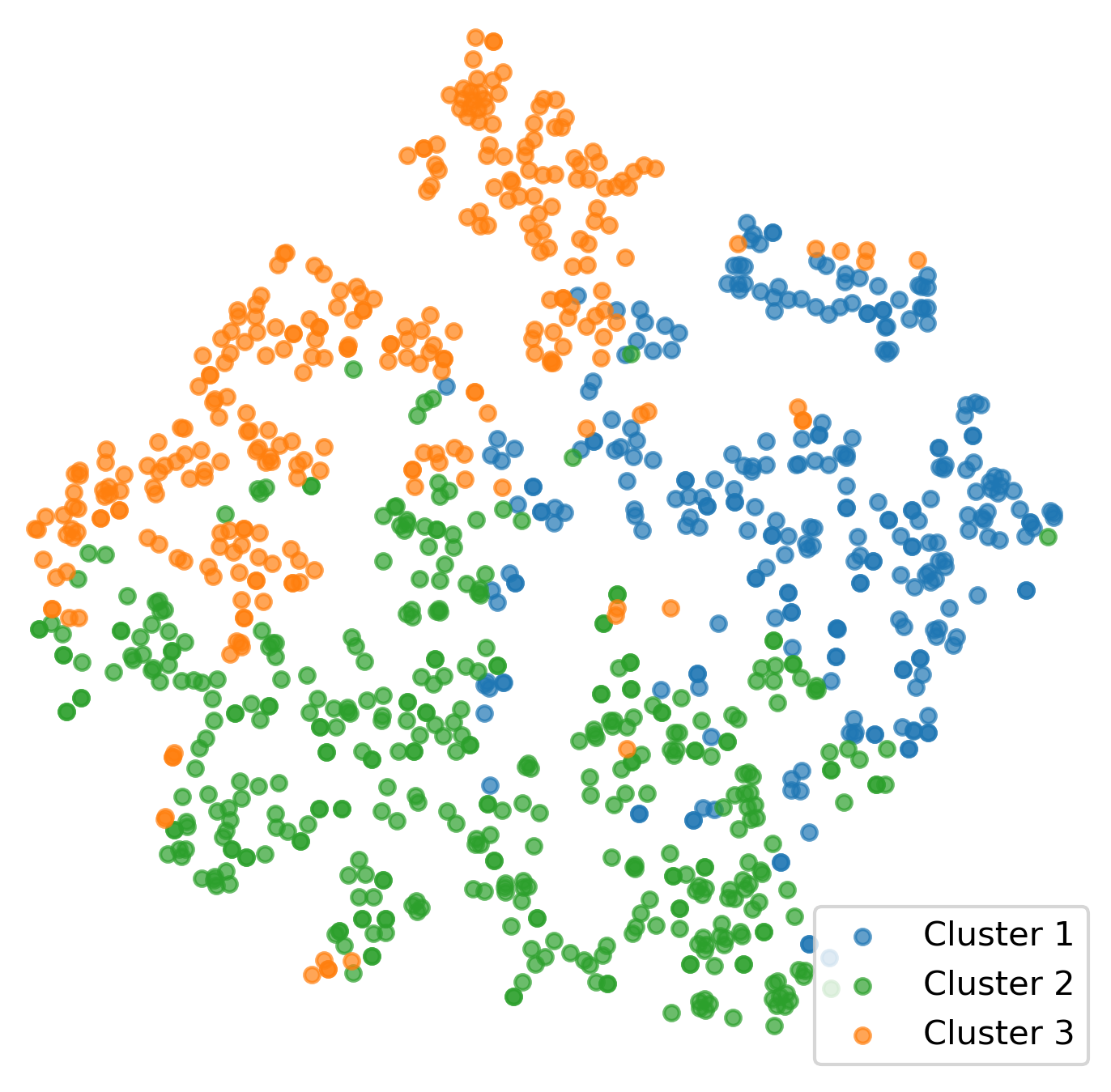}
        \captionsetup{justification=centering}
        \caption{VAE-GMM latent space color-coded using learned clusters} 
        \label{fig:vaegmm latent space cluster}
    \end{subfigure}
    \caption{The arrangement of level-type latent vectors (projected using t-SNE) of GMVAE and VAE-GMM.}
    \label{fig:latent space}
\end{figure}
}

\newcommand{\XFIGUREgenerations}{
\begin{figure*}[hbtp]
\centering
\begin{tabular}{cc}
\rotatebox{90}{\hspace{30mm}\textbf{SMB}} & \includegraphics[width=0.90\textwidth]{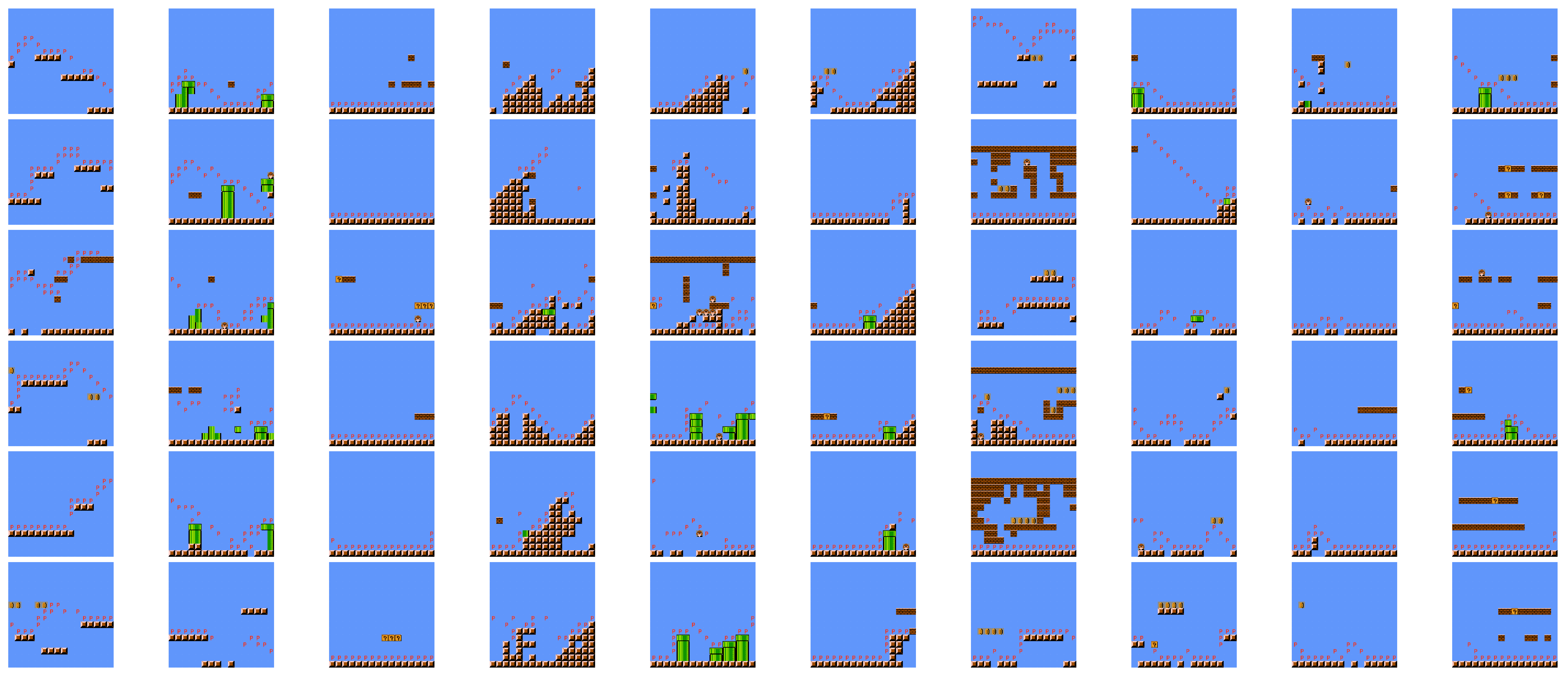} \\
\rotatebox{90}{\hspace{30mm}\textbf{KI}} & \includegraphics[width=0.90\textwidth]{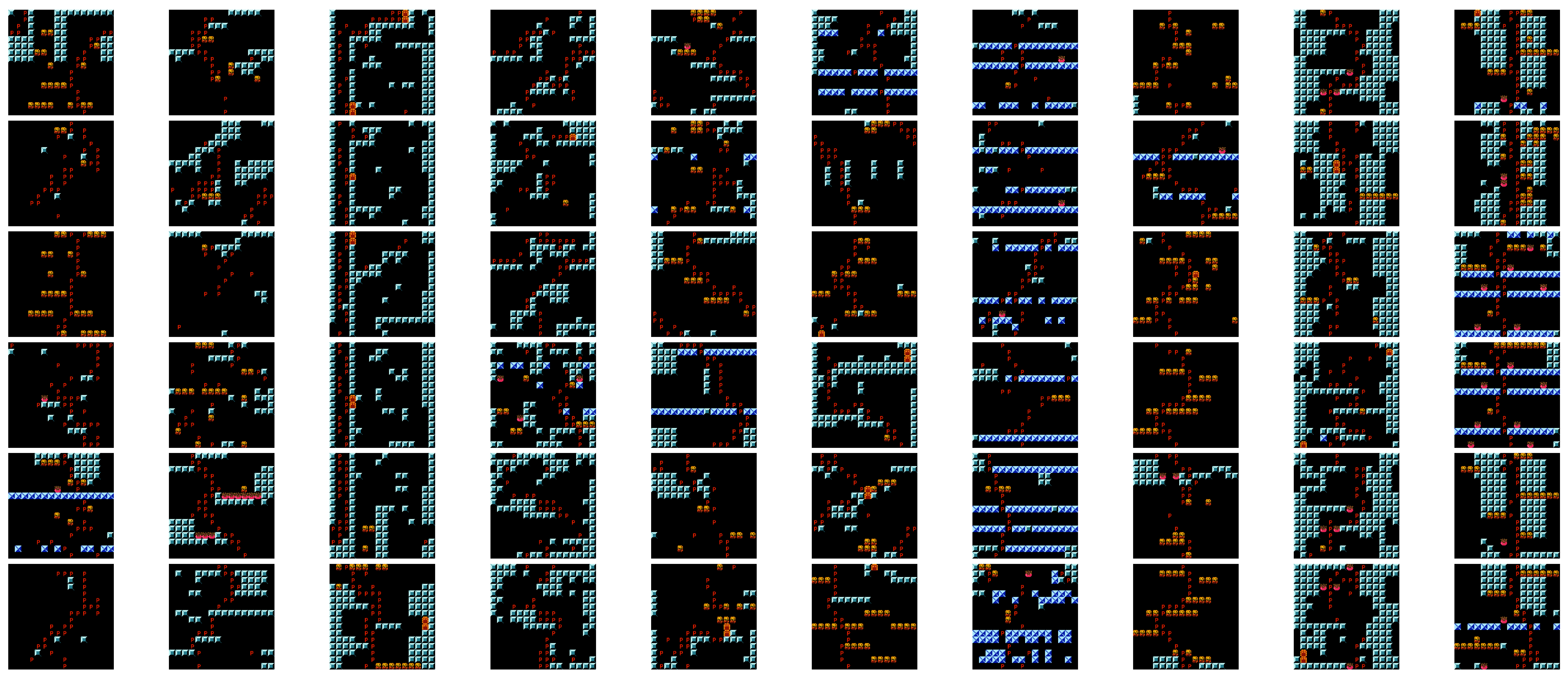} \\
\rotatebox{90}{\hspace{30mm}\textbf{MM}} & \includegraphics[width=0.90\textwidth]{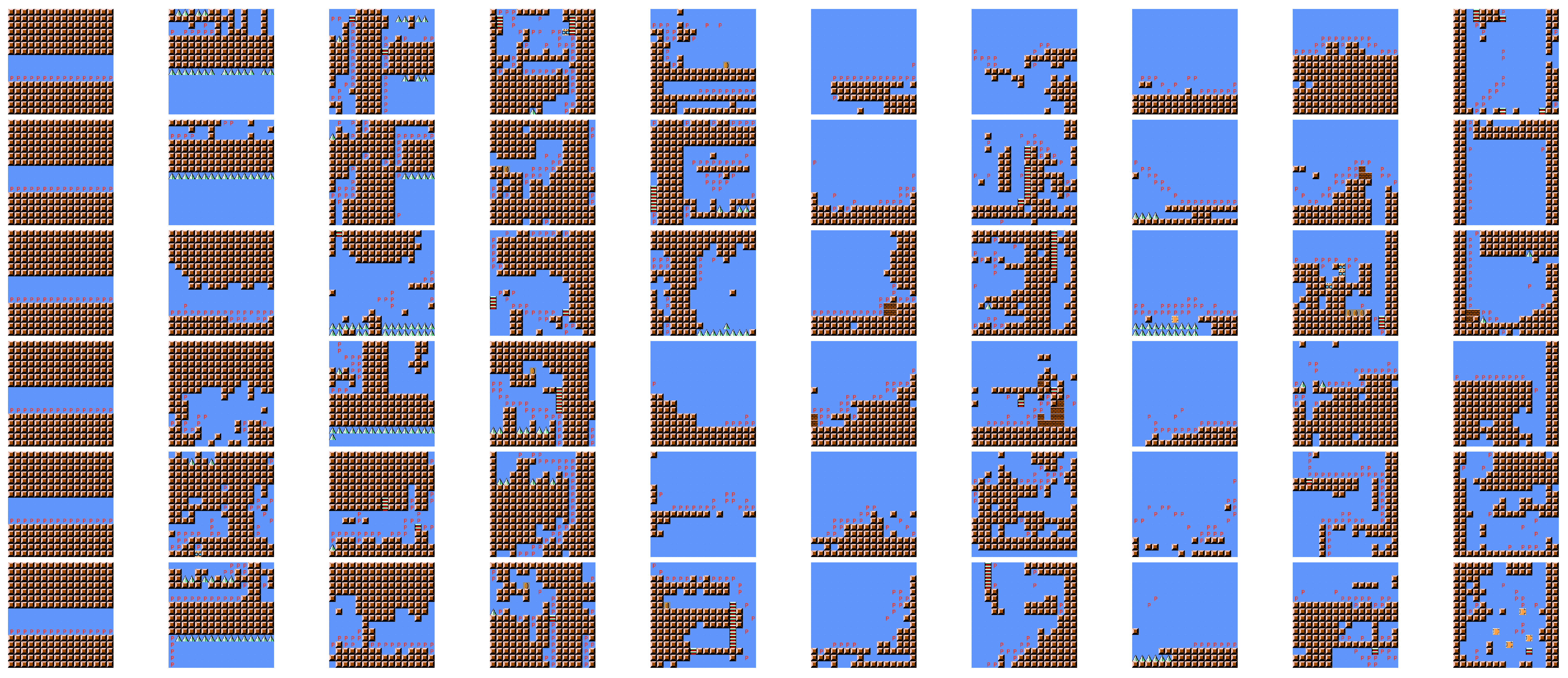} \\
& \includegraphics[width=0.90\textwidth]{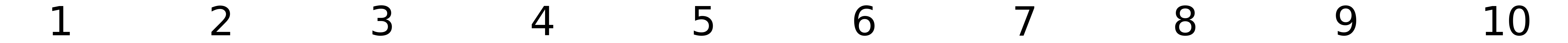}
\end{tabular}
\caption{Chunks generated by the 10-component GMVAE (one GMVAE trained on each game), with component indices shown at the bottom.}
\label{fig:generations}
\end{figure*}
}

\newcommand{\XFIGUREradialbarcharts}{
\begin{figure*}[!th]
\centering
\begin{tabular}{c}
\includegraphics[width=0.95\textwidth]{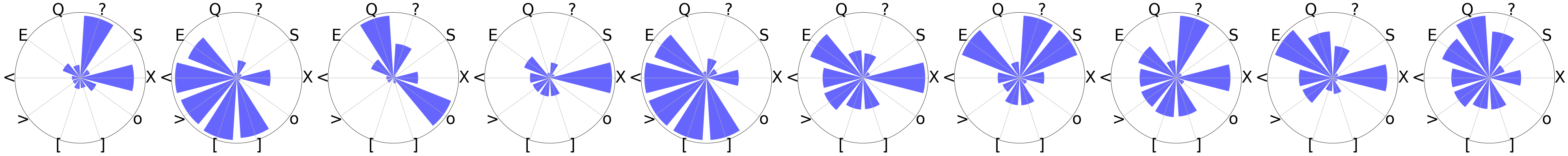} \\
\includegraphics[width=0.94\textwidth]{figure/component_indices}
\end{tabular}
\caption{Radial bar charts of mean tile densities of the 10-component GMVAE trained on SMB, with component indices shown at bottom. The mapping from characters to tiles: S (breakable), X (ground), o (coin), ] (lower right pipe), [ (lower left pipe), $>$ (upper right pipe), $<$ (upper left pipe), E (enemy), Q (empty question), ? (full question). - (background) has been excluded.}
\label{fig:radialbarcharts}
\end{figure*}
}

\newcommand{\XTABLEdisentanglementstats}{
\begin{table*}[t]
\centering
\scriptsize
\resizebox{\textwidth}{!}{
\begin{tabular}{|c|c|c|c|c|c|c|c|c|c|}
\hline
\multirow{2}{*}{p\% disentangled} &
  \multirow{5}{*}{\rotatebox{90}{\textbf{SMB}}} &
  GMVAE &
  VAE-GMM &
  \multirow{5}{*}{\rotatebox{90}{\textbf{KI}}} &
  GMVAE &
  VAE-GMM &
  \multirow{5}{*}{\rotatebox{90}{\textbf{MM}}} &
  GMVAE &
  VAE-GMM \\ \cline{3-4} \cline{6-7} \cline{9-10} 
                 &  & Mean Proportion & \# Components &  & Mean Proportion  & \# Components &  & Mean Proportion & \# Components \\ \cline{1-1} \cline{3-4} \cline{6-7} \cline{9-10} 
70\% (mediocre)  &  & 80.1\%          & 4            &  & 92.6\% & 4            &  & 78.0\%          & 6            \\ \cline{1-1} \cline{3-4} \cline{6-7} \cline{9-10} 
80\% (good)      &  & 61.0\%          & 4            &  & 74.4\% & 2            &  & 60.6\%          & 4            \\ \cline{1-1} \cline{3-4} \cline{6-7} \cline{9-10} 
90\% (excellent) &  & 13.1\%          & 2            &  & 16.2\% & 2            &  & 17.9\%          & 6            \\ \hline
\end{tabular}
}
\caption{Comparison of VAE-GMM and GMVAE using disentanglement statistics. For each game, we summarize the mean proportion of mediocre, good and excellent components for the GMVAE. We also state the number of components at which each proportion of the VAE-GMM drops below and never rises above the corresponding mean proportion of the GMVAE.}
\label{table:disentanglement stats}
\end{table*}
}

%================================================================================
% TABLES

%%%%%%%%%%%%%%%%%%%%%%%%%%%%%%%%%%%%%%%%%%%%%%%%%%
%% FIGURE/TABLE END %%%%%%%%%%%%%%%%%%%%%%%%%%%%%%
%%%%%%%%%%%%%%%%%%%%%%%%%%%%%%%%%%%%%%%%%%%%%%%%%%

%%%%%%%%%%%%%%%%%%%%%%%%%%%%%%%%%%%%%%%%%%%%%%%%%%
%% BODY BEGIN %%%%%%%%%%%%%%%%%%%%%%%%%%%%%%%%%%%%
%%%%%%%%%%%%%%%%%%%%%%%%%%%%%%%%%%%%%%%%%%%%%%%%%%

%================================================================================
\section{Introduction}
Variational autoencoders (VAEs) \cite{kingma2013autoencoding} have found use in games as an approach for Procedural Content Generation via Machine Learning (PCGML) \cite{summerville2018procedural} due to their ability to learn a continuous, latent representation of game level data which enables the generation of new levels via sampling and interpolation. However, not much PCGML work has leveraged the advantages of using the more advanced, hybrid variants of VAEs and related methods found in machine learning literature.
One such approach is the Gaussian Mixture Variational Autoencoder (GMVAE) developed by \citeauthoryearp{dilokthanakul2016deep}. Unlike regular VAEs which learn a latent space by imposing a Gaussian distribution as the prior, GMVAEs learn a latent space by imposing a prior of a mixture of Gaussians and allowing the means and variances of each component to be determined by training. Each component of the mixture thus learns to encode a meaningful subset of training data. When applied to game levels, such a model could help discover clusters of levels with similar characteristics. Moreover, each component could then be used as a generative model for producing levels with particular discovered features. While controllable level generation using latent models is also enabled by latent variable evolution \cite{bontrager2018deepmasterprints} as well as by other models such as conditional VAEs and GANs, the GMVAE has the potential of doing so in an unsupervised manner. 

Thus, in this paper, we train GMVAEs on chunks from \textit{Super Mario Bros.}, \textit{Kid Icarus} and \textit{Mega Man}. We motivate our approach by demonstrating that a GMVAE can cluster chunks from different level types in \textit{Super Mario Bros.} We then show that a GMVAE, with a reasonable number of components, can learn to cluster chunks by features such as the presence of specific game elements or design patterns with each component then being able to generate chunks from the corresponding cluster. Finally, we quantitatively confirm that GMVAEs cluster better than a baseline approach.

Our work thus contributes 1) a new PCGML approach for unsupervised clustering and generation of game levels of specific types and 2) to the best of our knowledge, the first application of GMVAEs, and Gaussian mixtures in general, for generating game levels.

\section{Background}

Methods for Procedural Content Generation via Machine Learning (PCGML) \cite{summerville2018procedural} attempt to generate game content using models trained on existing game data. Being a relatively new subfield of PCG research, most PCGML approaches thus far have focused on performing level generation via more well-known ML techniques such as Markov models \cite{snodgrass2017learning,dahlskog2014linear}, Bayes nets \cite{guzdial2016game,summerville2015samplinghyrule}, LSTMs \cite{summerville2016mariostring,sarkar2018blending}, autoencoders \cite{jain_autoencoders_2016}, GANs \cite{volz2018evolving,giacomello2018doom} and VAEs \cite{sarkar2019blending}. More advanced, hybrid ML models and architectures are not yet extensively used for PCGML. In this paper, we use such a model, the Gaussian Mixture VAE (GMVAE), for level generation.

GMVAEs were introduced by \citeauthoryearp{dilokthanakul2016deep} as a variant of the VAE \cite{kingma2013autoencoding} to be able to perform unsupervised clustering using latent variable modeling. While regular VAEs usually fix the prior distribution for the latent space to be a unimodal Gaussian, GMVAEs instead fix this prior to be a mixture of Gaussians. Each component of the mixture thus learns to encode a unique cluster of the training data. Further, the learned components can then be used to generate new data samples conforming to the features encoded by their respective cluster. This makes the GMVAE attractive for PCGML by building upon much of the work done using regular VAEs. \citeauthoryearp{sarkar2019blending} used VAEs to generate and blend levels of \textit{Super Mario Bros.} and \textit{Kid Icarus} while \citeauthoryearp{thakkar2019autoencoder} used VAEs to generate \textit{Lode Runner} levels. More recently, \citeauthoryearp{snodgrass2020multi} combined VAEs and binary space partitioning to create a multi-domain level generation approach using an abstract level representation. Outside of level generation, VAEs have been used to evolve controllers for \textit{Doom} \cite{alvernaz2017autoencoder} and classify NPC behavior \cite{soares2019deep}.

A number of prior works have utilized clustering for level generation. \citeauthoryearp{snodgrass2015hierarchical} used k-medoids clustering to discover level structures in \textit{Super Mario Bros.} and \textit{Lode Runner} for generating levels using multi-dimensional Markov chains. Similarly, \citeauthoryearp{guzdial2016learning} used k-means clustering to categorize learned level chunks of \textit{Super Mario Bros.} which were then used to train a probabilistic graphical model for generating new levels. Our approach differs in that our clusters are GMs in the latent space and are learned during the course of training the generative model, rather than fit to the input or output of a generative model as in the past works mentioned.

\XFIGUREgmvae

\section{Method}

\subsection{Dataset}

Our raw data consisted of the player-path-annotated, character-encoded levels from \textit{Super Mario Bros.} (SMB), \textit{Kid Icarus} (KI) and \textit{Mega Man} (MM) taken from the Video Game Level Corpus (VGLC) \cite{summerville2016vglc}. For each level, we obtained 16x16 chunks by sliding a 16x16 window across the level one tile at a time. We ended up with 2698 SMB chunks, 1142 KI chunks and 3330 MM chunks. We then used integers to encode the chunks. For training, we converted the integer encoding into one-hot encoding and flattened the tensor representing each chunk as a 1-dimensional vector. For visualizing some of the MM and KI levels, we use a mixture of sprites from those games as well as SMB. Paths are visualized using the letter 'P'.

\subsection{GMVAE Architecture}
In a VAE \cite{kingma2013autoencoding} with an $l$-dimensional latent space, the encoder takes in an input vector and outputs $l$ pairs of means and variances that parameterize the input latent distribution corresponding to that input vector. A latent vector is then sampled from this distribution and forwarded through the decoder which outputs a reconstruction of the input vector. Two loss terms are minimized: (1) the binary cross-entropy loss between the input and output vectors and (2) the KL-divergence between the input latent distribution and the Gaussian prior. A GMVAE with a $k$-component GM prior (Figure \ref{fig:gmvae}) makes the following two modifications to such a VAE.

\textbf{First modification} The input vector is first passed through a \textit{label-assigning} network, whose last layer, the Gumbel-Softmax layer, produces a $k$-dimensional label. Its $i$-th dimension contains the probability that the input vector belongs to the $i$-th GM component. During training, this set of probabilities is gradually enforced to be concentrated on one component \cite{jang2016categorical}. Therefore, during generation and later stages in training, this layer outputs one-hot labels. The concatenation of the input vector and this label is fed into the encoder.  

\textbf{Second modification} The second loss term is changed. For each input vector, we now minimize the KL-divergence between its input latent distribution and its assigned \textit{component latent distribution}, which is parameterized by the \textit{component means} and \textit{component variances} obtained by forwarding its label through the \textit{prior-assigning} network. This new loss term is called the \textit{GM-prior loss}.

\subsection{GMVAE Modules}

Let $d$ be the dimension of input and output vectors (reconstructions), $k$ be the number of components of the GM prior and 64 be the latent space dimension. Here, $d$ is game-specific: 3072 (16 width $\times$16 height $\times$12 unique tiles) for SMB, 1792 (16$\times$16$\times$7) for KI and 4352 (16$\times$16$\times$17) for MM. Then, the label-assigning network has an input dimension of $d$ and contains 3 fully-connected layers using 512 neurons per layer and ReLU activations, plus a GumbelSoftmax layer using $k$ neurons.

The prior-assigning network has two independent sub-networks. The first has an input dimension of $k$ and contains 1 fully-connected layer using 64 neurons and a linear activation. Its purpose is to compute the means of the GM components assigned by the label-assigning network. The second sub-network is the same as the first except that it uses the Softplus activation. Its purpose is to compute the variances of the GM components assigned by the label-assigning network. The number of neurons in these two sub-networks corresponds to the fact that an isotropic GM in a 64-dimensional latent space has 64 means and variances.

The encoder has two connected sub-networks. The first has an input dimension of $d+k$ and contains 3 fully-connected layers using 512 neurons per layer and ReLU activations. The second contains 2 independent fully-connected layers: one using 64 neurons and a linear activation and another using 64 neurons and a Softplus activation. The second sub-network takes in the output of the first sub-network and computes the means and variances of the latent distributions.

The decoder network has an input dimension of 64 and contains 3 fully-connected layers using 512 neurons per layer and ReLU activations plus an output layer using $d$ neurons and a Sigmoid activation.

\subsection{Training GMVAE}

Each GMVAE was trained using a batch size of 64 for 10000 epochs with the Adam optimizer at a learning rate of 0.001. The scalar weights on the GM-prior loss and on the reconstruction loss were 2 and 1 respectively. All models were trained using PyTorch \cite{paszke2017automatic}. 

\subsection{Generating from GMVAE}

If a component clusters chunks with a specific design pattern, then chunks of the same style can be generated from that component. To generate $n$ chunks from the $i$-th GM component of a $k$-component GMVAE, we first create a one-dimensional array of $n$ component indices $i$. This array is then one-hot encoded and fed into the prior-assigning network to get the parameters of the $i$-th GM component. Finally, $n$ latent vectors are sampled from this component and fed into the decoder to obtain generated chunks. Since these generations are one-hot encoded, an argmax operation on the one-hot dimension is performed to obtain the integer-encoded chunks.

\XFIGUREgenerations
\XFIGUREradialbarcharts

\XFIGURElatentspacevis

\section{Results and Discussion}

\subsection{Visual inspection of GMVAE generations}

Our goal of applying the GMVAE is to have each GM component cluster and generate chunks with similar characteristics. Since this is done in an unsupervised manner, the primary utility of this approach is the discovery of patterns and structures in the levels of a game. Thus, while a designer would not be able to explicitly control what properties the components learn to cluster, such an approach would be useful for helping designers find new structures and patterns they hadn't considered. The learned components can then be examined by a simple visual inspection of levels generated using each component as in Figure \ref{fig:generations}. Due to limited space, we included only six generations per component.

For a more robust evaluation, we also consider radial bar charts of mean tile densities for each component (Figure \ref{fig:radialbarcharts}). Note that the mean densities of a specific tile (one value for each component) are divided by their maximum value for normalization. Such charts differentiate between tiles by placing them at different angles. Doing so signifies the fact that the clusters are different because they have ``protrusions'' (bars of high-density tiles) at different angles.

To see how a human designer might use radial bar charts to better understand the components of a GMVAE, we consider the example of SMB. In the radial bar charts for SMB (Figure \ref{fig:radialbarcharts}), we can see that component 4 has a very large value of ``X'' relative to other components, which corresponds to ground tiles in the VGLC. Note that this observation requires no comparison between component 4 and other components because the density values are normalized by tile. Indeed, in Figure \ref{fig:generations}, for SMB, we see that component 4 clusters the staircase design pattern, which uses a large number of ground tiles. Component 5 and 6 also cluster this design pattern, but they are less consistent. This can be seen from both Figure \ref{fig:generations} and the fact that their radial bar charts also protrude at other tiles (Figure \ref{fig:radialbarcharts}).

\subsection{Advantages of GMVAE over VAE and Gaussian Mixture Model (GMM)}

One may be interested in comparing a) fitting a GMVAE and b) first fitting a VAE and then fitting a GMM to its latent space. We refer to the latter hybrid approach as VAE-GMM. 

First, consider the advantage of fitting a GMVAE. During the training of a GMVAE, latent vectors are encouraged to gradually be distributed closer to their assigned GM components. However, doing so alone does not guarantee that components are assigned based on similarity between chunks and can result in clusters that are not meaningful. This is addressed by the fact that the learned component labels are also passed into the encoder as additional information helpful for reconstruction. Intuitively, the labels for dissimilar chunks need to be different in order to inform later parts of the network to reconstruct certain aspects differently; conversely, the labels for similar chunks should be identical.

On the other hand, the arrangement of latent vectors in the latent space of a VAE is completely determined by the KL-divergence with a Gaussian prior and the reconstruction loss. These two terms do not explicitly account for clustering. In fact, in being unimodal, the Gaussian prior may be detrimental to clustering. Thus, we hypothesize that meaningful clusters might be hard to extract from a VAE's latent space using conventional clustering techniques like GMM. If this is true, then GMVAE would have a significant advantage over this naive post-fitting approach. We verify this hypothesis through two experiments. 

\textbf{First experiment} In the first experiment, we trained a 3-component GMVAE and a 3-component VAE-GMM on the \textit{SMB} data. In SMB, there are 3 non-overlapping level types: overworld, underworld and jumpy levels with each chunk belonging to exactly one of these types. Overworld levels are levels above ground; underworld levels are levels below ground and are characterized by having rows of ground tiles at the top of each chunk; jumpy levels consist of floating platforms without ground tiles at the bottom. The goal is to see which model better cluster these three level types in its latent space. A weighted random sampler was used during training to mitigate the imbalance among the three level types and was also used in the final forward pass of all SMB input vectors to obtain the latent vectors.

The 3-component VAE-GMM was obtained by 1) training a standard VAE, 2) obtaining the latent vectors by feeding the input vectors into the encoder, 3) applying Principle Component dimensionality reduction (keeping 95\% variance) to the latent vectors and 4) fitting a 3-component GMM on the projected latent vectors. Dimensionality reduction was applied because GMMs are known to have difficulty converging in high-dimensional spaces.

The GMVAE and VAE-GMM achieved 88.5\% and 63.3\% clustering accuracy respectively after balancing level types. To better understand these results, we used t-SNE to visualize the arrangement of latent vectors of GMVAE and VAE (Figure \ref{fig:latent space}) as t-SNE is a cluster-preserving algorithm. The GMVAE arranges the latent vectors in 3 well-separated clusters (Figure \ref{fig:gmvae latent space cluster}), with each cluster corresponding to a unique level type (Figure \ref{fig:gmvae latent space label}). For the VAE, while the latent vectors from the same level type distribute closer, the cluster boundaries are not well-separated (Figure \ref{fig:vaegmm latent space label}), making them difficult to be extracted by the GMM as hypothesized (Figure \ref{fig:vaegmm latent space cluster}).

\XFIGUREdisentanglement
\XTABLEdisentanglementstats

\textbf{Second experiment} This is a more comprehensive version of the first experiment. Here, we trained GMVAEs and VAE-GMMs with 2, 4, 6, 8, 10, 15, 20, 30, 40 and 50 components on the \textit{SMB} data, the \textit{KI} data and the \textit{MM} data. Again, we want to compare the clustering performance of these 2 models, but in more settings. However, unlike the first experiment, since we did not know what clusters to expect, we did not have true labels using which we could compute clustering accuracy. Instead, we proposed a different metric.

For each model, we generated 500 chunks (300 for training; 200 for validation) from each component. Using these chunks and their component labels, we trained a multi-layer perceptron (MLP) classifier. We defined a component as $p\%$-\textit{disentangled} from other components if the MLP achieves $p\%$ accuracy on the validation set of that component.

We refer to components that are ${\geq} 70\%$, ${\geq}80\%$ and ${\geq} 90\%$ disentangled as mediocre, good and excellent respectively. Figure \ref{fig:disentanglement} shows the proportion of such components against the total number of components. These proportions drop drastically as the number of components increases for the VAE-GMM but are relatively high and stable for the GMVAE. More detailed statistics are available in Table \ref{table:disentanglement stats}. We postulate that this is because GMVAE can automatically adjust the granularity of design patterns clustered by each component. Nevertheless, the GMVAE performs poorly at 2 components for both SMB and MM. This is likely because having only 2 components causes multiple design patterns to be clustered by each component, making the component labels less discriminating and less informative for reconstruction, and are therefore poorly learned during training. 

\subsection{Playability of GMVAE generations}

Although playability is not the focus of this paper, it is an important quality measure for level generation. To evaluate the playability of a $k$-component GMVAE, we first sample $\textrm{floor}(\frac{10000}{k})$ chunks from each component and aggregate them. Then we apply the A* search code available from the VGLC which we have adapted to game chunks. In experiment 2, all GMVAEs trained on SMB achieved at least 99.4\% playability and those trained on KI achieved at least 79.2\%. Although these scores might be satisfactory for chunks, the playability of whole levels made up of such chunks could drop as more chunks are generated. We did not evaluate playability for MM chunks because, unlike SMB and KI which contain only horizontal and only vertical levels respectively, MM levels use a mixture of these two traversal directions, making it ambiguous to set the starting and finishing criteria when adapting the A* search code to chunks. 

\section{Conclusion and Future Work}
We demonstrated that GMVAEs are able to cluster game chunks with specific design patterns in an unsupervised manner and subsequently generate new chunks from each cluster. In addition, we showed that GMVAEs are better at clustering than a naive VAE-GMM approach, by contrasting their latent spaces and disentanglement proportions. 

The GMVAE can be a good exploratory tool that helps understand the variety of designs present in a game. However, its unsupervised nature leads to some practical limitations. First, it takes careful inspection of generations and summary statistics to fully comprehend the design pattern(s) clustered by each component. Second, while the components are shown to be highly disentangled, each component may cluster multiple design patterns in which case we cannot exclusively generate one specific pattern from a component. Third, it is unclear whether GMVAE clusters in a way similar to how humans would. Thus, in the future, we could conduct user studies to determine how these clusters could be used by designers and how such a model could help designers find new structures and patterns.

Future work could also look into whether the GMVAE inherits desirable properties of the VAE, such as smooth interpolation between two chunks when interpolating their latent vectors. Given Figure \ref{fig:gmvae latent space cluster}, one may hypothesize that interpolation between chunks in separate clusters is no longer useful since the interpolated vectors might fall in the space between clusters. However this space may also have potential for novelty. For e.g., could we find new structures, patterns and level types in the space between learned clusters?

Our approach generates level chunks rather than whole levels, similar to past works using latent models, due to the low number of full training levels and because such models work with fixed-size inputs and outputs. A naive way of generating whole levels is iteratively generating, filtering and concatenating successive chunks but this does not guarantee playability since chunks are generated independently. To this end, \citeauthoryearp{sarkar2020sequential} presented a sequential latent model of level generation that produces whole levels consisting of segments following a logical progression. Further, autoregressive models like LSTMs have been shown to generate playable levels of arbitrary length \cite{summerville2016mariostring}. Thus in the future we could explore approaches that combine GMVAEs with such models.

%%%%%%%%%%%%%%%%%%%%%%%%%%%%%%%%%%%%%%%%%%%%%%%%%%
%% BODY END %%%%%%%%%%%%%%%%%%%%%%%%%%%%%%%%%%%%%%
%%%%%%%%%%%%%%%%%%%%%%%%%%%%%%%%%%%%%%%%%%%%%%%%%%

% References and End of Paper
% These lines must be placed at the end of your paper
\bibliography{refs-manual}
\bibliographystyle{aaai}
\end{document}